\pdfoutput=1

\documentclass[11pt]{article}

\usepackage[preprint]{acl}

\usepackage{times}
\usepackage{latexsym}

\usepackage[T1]{fontenc}

\usepackage[utf8]{inputenc}

\usepackage{microtype}

\usepackage{inconsolata}

\usepackage{graphicx}
\usepackage{amsmath} 
\usepackage{verbatim}
\usepackage{makecell}
\usepackage{multirow}
\usepackage{array}
\usepackage{booktabs}

\title{Domain-specific Guided Summarization for Mental Health Posts}

\author{\textbf{Lu Qian\textsuperscript{1,2}}, \textbf{Yuqi Wang\textsuperscript{1,2}}, \textbf{Zimu Wang\textsuperscript{1,2}}, \textbf{Haiyang Zhang\textsuperscript{1}}, \\
\textbf{Wei Wang\textsuperscript{1,\thanks{Corresponding author.}}}, \textbf{Ting Yu\textsuperscript{3}}, \textbf{Anh Nguyen\textsuperscript{2}} \\
\textsuperscript{1}School of Advanced Technology, Xi’an Jiaotong-Liverpool University, China \\
\textsuperscript{2}Department of Computer Science, University of Liverpool, UK \\
\textsuperscript{3}School of Information Science and Technology, Hangzhou Normal University, China \\
\small\texttt{\{Lu.Qian21,Yuqi.Wang17,Zimu.Wang19\}@student.xjtlu.edu.cn} \\
\small\texttt{\{Haiyang.Zhang,Wei.Wang03\}@xjtlu.edu.cn, yut@hznu.edu.cn, Anh.Nguyen@liverpool.ac.uk}}

\begin{document}
\maketitle
\begin{abstract}
In domain-specific contexts, particularly mental health, abstractive summarization requires advanced techniques adept at handling specialized content to generate domain-relevant and faithful summaries. 
In response to this, we introduce a guided summarizer equipped with a dual-encoder and an adapted decoder that utilizes novel domain-specific guidance signals, i.e., mental health terminologies and contextually rich sentences from the source document, to enhance its capacity to align closely with the content and context of guidance, thereby generating a domain-relevant summary. 
Additionally, we present a post-editing correction model to rectify errors in the generated summary, thus enhancing its consistency with the original content in detail.
Evaluation on the \textsc{MentSum} dataset reveals that our model outperforms existing baseline models in terms of both \textsc{Rouge} and FactCC scores. 
Although the experiments are specifically designed for mental health posts,
the methodology we've developed offers broad applicability, highlighting its versatility and effectiveness in producing high-quality domain-specific summaries.
\end{abstract}

\section{Introduction}

Mental health is a critical area that profoundly affects both individuals and society, demanding effective and accurate communication for support \cite{hua2024large}. In this domain, abstractive summarization plays a pivotal role by condensing one lengthy user post from online platforms like Reddit\footnote{\url{https://www.reddit.com}} and Reachout\footnote{\url{https://au.reachout.com}} into a concise summary. This process, through paraphrasing, generalizing, and reorganizing content with novel phrases and sentences, effectively conveys the essential information and meaning of the original text \citep{Shi2021, Qian2023}.
The summary enables quicker review and response by professional counselors, thus enhancing support for individuals dealing with mental health issues and demonstrating significant social impact.

\begin{figure}[t!]
    \centering{\includegraphics[width=0.48\textwidth]{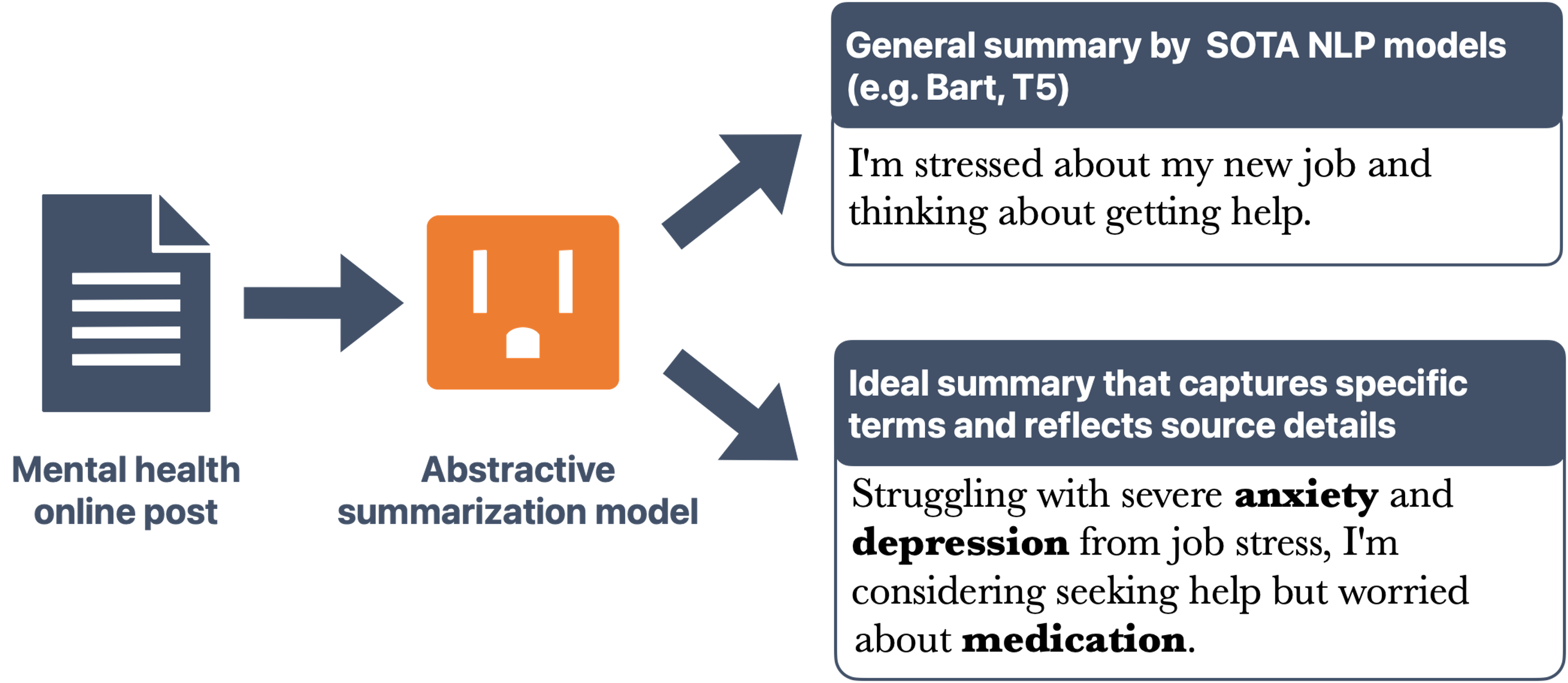}}
    \caption{This example highlights the importance of an ideal summary that, compared to a general summary, is focused on domain relevance and faithful to the source post, providing essential support for effective communication within the mental health community.}
    \label{fig}
\end{figure}

Despite advancements in natural language processing (NLP), applying abstractive summarization to mental health posts illustrates some major challenges in domain-specific summarization.
The first challenge is that the summary generated by state-of-the-art (SOTA) pre-trained models \citep{liu-lapata-2019-text, lewis-etal-2020-bart, Raffel2020} tends to be too general and \textit{lacks domain specificity}.
These models often struggle to control the content of the summary, making it difficult to determine in advance which parts of the original content should be emphasized \citep{dou-etal-2021-gsum}. 
The second challenge pertains to the \textit{faithfulness} of the generated summary. Often, there is a notable risk of producing a summary that may contradict or diverge from the source document, potentially introducing intrinsic hallucination\footnote{Intrinsic hallucination refers to content in a generated summary that contradicts the source document.} or inconsistency 
\citep{kryscinski-etal-2020-evaluating, wang2024knowledge, wang2024generating, na2024rethinking}. 
Together, these issues highlight the need for more advanced summarization techniques that can adeptly handle the complexities of domain-specific content while ensuring contextual relevance and detail consistency, as shown in Figure \ref{fig}.

Drawing inspiration from the \textsc{GSum} \citep{dou-etal-2021-gsum} framework for its ability to enhance controllability through guidance signal and constrain summary to deviate less from the source document, we introduce a guided summarizer featuring a dual-encoder and an adapted decoder architecture that leverages two types of domain-specific knowledge-based guidance, i.e., specialized mental health terminologies and contextually rich sentences from source post. 
This design is specifically tailored to enhance the summarization process within mental health contexts, guiding the generation of a summary that is both terminologically precise and richly informed by the underlying domain-specific information contained within the original text.

Further, building on established post-editing practice in recent studies \citep{dong-etal-2020-multi,cao-etal-2020-factual}, we propose a corrector that follows the summarizer and is dedicated to identifying and correcting potential inconsistencies in the generated summary with respect to the source post.
This step ensures the corrected summary more faithfully represents the details of the original text.
At last, we evaluate our model on \textsc{MentSum} \citep{sotudeh-etal-2022-mentsum}, the first mental health summarization dataset. The output summary is evaluated by not only the \textsc{Rouge} scores \citep{lin-2004-rouge} measuring linguistic quality, but also FactCC score \citep{kryscinski-etal-2020-evaluating}, an automatic metric assessing factual consistency\footnote{Although recent studies define ``factuality” as being based on real-world facts, our paper uses the term ``factual consistency”, which is commonly employed in evaluation research, to emphasize alignment with the source document.} with the source document. 

The contributions of this study are as follows: 
\begin{itemize}
    \item We introduce novel domain-specific guidance signals, encoded by a separate encoder to guide the summarization process to align closely with the content and context of guidance, thus improving the summary's domain relevance.
    \item We propose a correction model as a subsequent enhancement step to identify and rectify any potential inconsistency in the generated summary, thereby reducing intrinsic hallucination and further improving faithfulness.
    \item Our top-performing model, using contextually rich sentences as guidance, outperforms the previous SOTA model \textsc{CurrSum} \citep{sotudeh-etal-2022-curriculum}, achieving improvements of 0.40, 0.82, and 4.07 in \textsc{Rouge-1}, \textsc{Rouge-2}, and \textsc{Rouge-L} scores, respectively. Furthermore, it achieves a 2.5\% higher FactCC score compared to \textsc{Bart}, and a 3.0\% increase over the original \textsc{GSum}.
\end{itemize}

\begin{figure*}[t!]
\centerline{\includegraphics[width=1.0\textwidth]{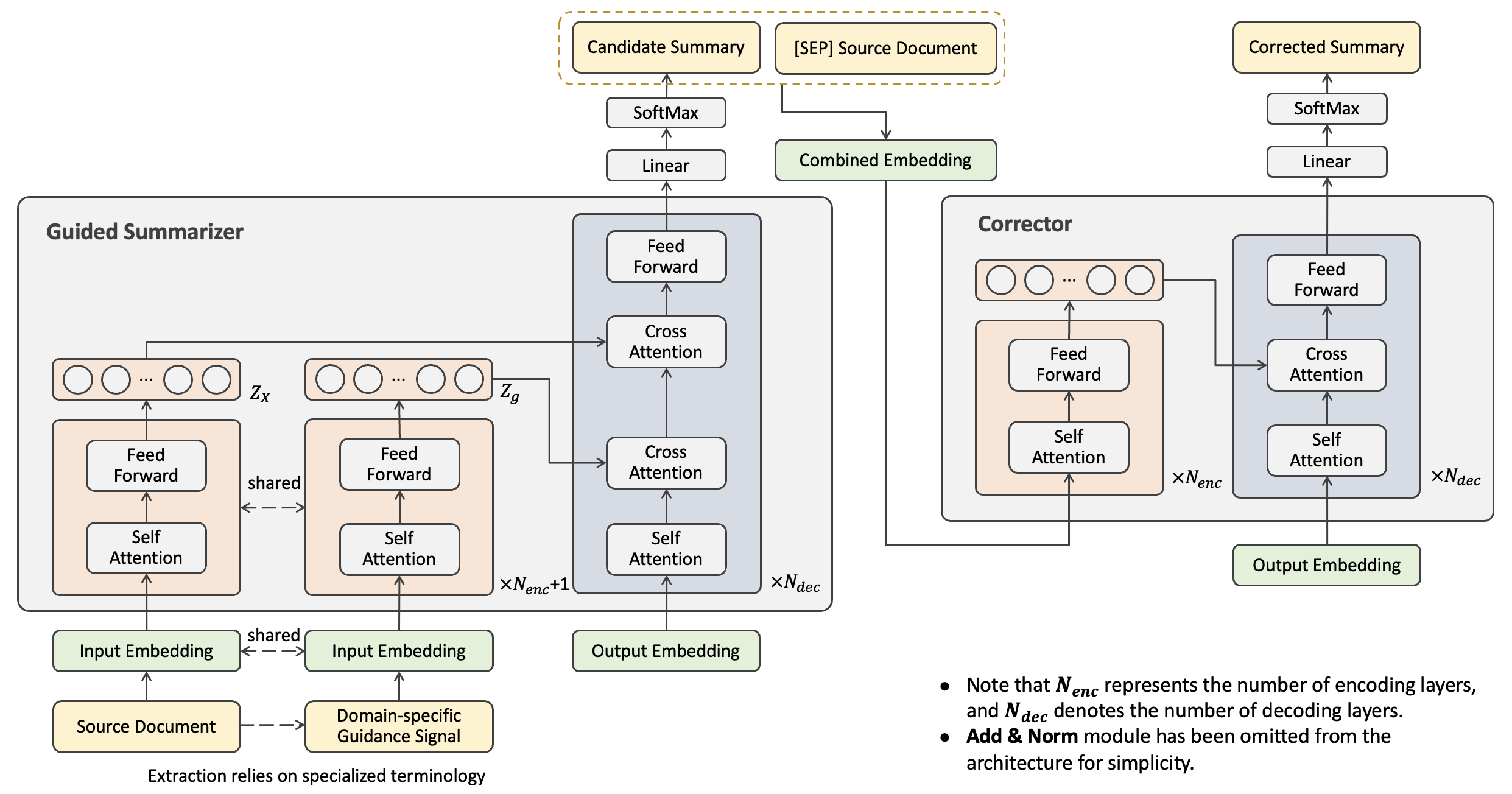}}
\caption{The overall architecture: The initial phase involves a guided summarizer with a dual-encoder and an adapted decoder architecture, utilizing domain-specific guidance signals to produce a candidate summary. This is then refined in the second phase by a post-editing corrector, which identifies and corrects potential inconsistencies in the candidate summary with respect to the source document.}
\label{framework}
\end{figure*}

\section{Related Work}
\subsection{Guided Abstractive Summarization} \label{limit}
The development of neural abstractive summarization has seen significant advancements through the implementation of sequence-to-sequence (seq2seq) framework \citep{chopra-etal-2016-abstractive, nallapati-etal-2016-abstractive1} and the Transformer architecture \citep{Vaswani2017, lewis-etal-2020-bart, Raffel2020}. Building on these foundations, guided abstractive summarization leverages additional guidance signals or user input to steer the summarization process, ensuring that the resulting summary is aligned with the specific need and preference. 

Knowledge bases (KBs) are the most popular guidance and enable summarization systems to deeply engage with the semantic relationship and hierarchical structure they encapsulate. 
Internal KBs \citep{huang-etal-2020-knowledge, zhu-etal-2021-enhancing} extract knowledge directly from source documents using information extraction tools \cite{wang2024document}, reducing intrinsic hallucination and improving the summary's faithfulness. Meanwhile, external KBs \citep{Liu2021c, dong-etal-2022-faithful, Zhu2024} provide common-sense or world knowledge, enhancing the factuality and reliability of the generated summary. 

For other guidance, \citet{he-etal-2022-ctrlsum} and \citet{narayan-etal-2021-planning} incorporate user-defined keywords and learned entity prompts, respectively.
Moreover, \citet{dou-etal-2021-gsum} expands on these ideas with the \textsc{GSum} framework, which supports different types of guidance signals, i.e., highlighted sentences, keywords, salient relational triples, and retrieved summaries. 

\subsection{Domain-specific Summarization}
Domain-specific summarization, particularly in the healthcare field, faces challenges due to the complexity of terminology, the critical need for accuracy in health-related decisions, and the concern over patient confidentiality and data privacy. However, the emergence of advanced NLP techniques and the availability of large annotated medical datasets have spurred increased interest and progress in this area.

Key efforts include the development of automated radiology report summarization to help streamline healthcare by turning complex radiographic findings into concise summaries, supported by datasets like Indiana University chest X-ray collection (OpenI) \citep{DemnerFushman2015} and MIMIC-CXR \citep{Johnson2019}. Similarly, innovative approaches like the Re$^3$Writer model \citep{Liu2022b} leverages the ``Patient Instruction” (PI) dataset from MIMIC-III to generate discharge instructions tailored to individual patient records by simulating the physician decision-making process. 
Additionally, efforts to summarize varied hospital course notes into Brief Hospital Course (BHC) summaries \citep{Searle2023} utilize adapted \textsc{Bart} model, enhanced with clinical ontology signals for producing problem-list-orientated summaries. 
Furthermore, the creation of the \textsc{MentSum} \citep{sotudeh-etal-2022-mentsum} dataset for mental health online posts summarization on Reddit further exemplifies the domain's growing research interest, with models like \textsc{CurrSum} employing curriculum learning strategy to improve performance. 
These advancements highlight the evolving landscape of healthcare summarization, driven by a blend of the latest NLP technologies and domain-specific knowledge.

\section{Methodology}

The overall architecture of our proposed model is illustrated in Figure \ref{framework}. 
By leveraging the strength of both guided summarization and correction in a unified framework, this integrated approach aims to generate summaries that are both domain-relevant and faithful, addressing the challenges of domain-specific summarization.

\subsection{Guided Summarizer}

\paragraph{Domain-specific Guidance Signal.}
The core innovation of our model lies in introducing domain-specific guidance signals, encoded by a separate encoder and designed to steer the summarization process to closely align with the content and context of guidance. 
Specifically, we extract two types of guidance signals from source posts: specialized mental health terminologies and, separately, sentences that contain any of these identified terms.
Intuitively, incorporating this knowledge-based guidance would help the summary enhance domain specificity by adhering to specialized terminologies and emphasizing relevant underlying information within the original text \citep{wang2023fusing}. 
More details about the guidance extraction are described in Section \ref{details}. 

\paragraph{Dual-encoders.}
The first encoder transforms source document $X = (x_1, ..., x_n)$ into a sequence of contextual representations $Z_X = (z_{x_1}, ..., z_{x_n})$, while the second encoder processes domain-specific guidance signal $g = (g_1, ..., g_k)$, which can be either terms or sentences, into a sequence of guidance representations $Z_g = (z_{g_1}, ..., z_{g_k})$, where $k$ is the length of the guidance input. Employing self-attention and feed-forward blocks followed by layer normalization, each encoder yields the output vector that encapsulates rich contextual and guidance-driven information for each token in both the document and the guidance.

\paragraph{Decoder.}
The decoder then integrates outputs from both encoders to generate the summary $Y = (y_1, ..., y_m)$. Modifications have been made to the standard Transformer's decoder structure, enabling it to attend to both the document and the guidance, instead of just one input sequence. Specifically, in each decoding layer, after the self-attention block, the decoder first attends to the guidance representations $Z_g$, enabling it to decide which part of the source document should be focused on. Then, it uses these signal-aware intermediate representations to more effectively attend to the document representations $Z_X$, culminating in a summary that is both informative and aligned with the guidance. 

\paragraph{Training Objective.}
The objective function aims to maximize the log-likelihood of generating the summary $Y$ given both the source document $X$ and the guidance signal $g$. It is formulated as:
\begin{equation}
\begin{split}
\arg \max_\theta \sum_{i=1}^{N} \log P(Y^{(i)}|X^{(i)}, g^{(i)}; \theta) \\
= \arg \max_\theta \sum_{i=1}^{N} \sum_{t=1}^{m^{(i)}} \log P(y_t^{(i)}|y_{<t}^{(i)}, X^{(i)}, g^{(i)}; \theta),
\end{split}
\end{equation}
where $N$ is the number of training examples, $Y^{(i)}$, $X^{(i)}$, and $g^{(i)}$ represent the summary, source document, and guidance for the $i$-th example, respectively, and $\theta$ denotes the learnable parameters of our model. This can be further decomposed into the sum of the log probabilities of each token in the summary conditioned on the preceding tokens, the source document, and the guidance, where $m^{(i)}$ is the length of the $i$-th summary, and $y_{<t}^{(i)}$ denotes all generated tokens in the $i$-th summary before position $t$.

By optimizing this function, our model learns to produce one summary that not only captures the essence of the source document but also closely adheres to the guidance signal. 
During training, the parameters of the word embedding layers and the bottom encoding layers are shared between the two encoders to reduce the computation and memory requirements, while the top layers of the two encoders are distinct, and initialized with pre-trained parameters but separately trained for each encoder. In the decoder, the first cross-attention block is initialized randomly since it is additional to the standard Transformer structure, while the second cross-attention block is initialized with pre-trained parameters.

\subsection{Corrector}
In addition to the guided summarizer, we propose a neural corrector as a subsequent enhancement to identify and rectify potential inconsistencies in the generated summary with respect to the source document.
This correction process can be modeled as a seq2seq problem: given a candidate summary $Y$ and its corresponding document $X$, it aims to produce a corrected summary $Y'$ that is more consistent with the original document $X$.

\paragraph{Artificial Corruption Data.}
To adequately train the neural corrector, we generate synthetic examples by introducing intentional errors based on heuristics by \citet{kryscinski-etal-2020-evaluating}. This involves creating incorrect summaries by swapping entities, numbers, dates, or pronouns using a strategy outlined by \citet{cao-etal-2020-factual}. Specifically, the first three swaps are made by replacing one item in the reference summary with another random item of the same type from the source document, while the pronoun swap is made by replacing one pronoun with another one of a matching syntactic case. 

\paragraph{Model Design.}
The correction model is designed to rectify an incorrect summary $Y$ into a consistent summary $Y'$ with minimal modifications based on the source document $X$. This can be formulated as optimizing the model parameters $\theta$ to maximize the likelihood function within an encoder-decoder framework: 
\begin{equation}
\arg \max_\theta \sum_{i=1}^{N} \log P(Y'^{(i)}|Y^{(i)}, X^{(i)}; \theta),
\end{equation}
where $N$ is the number of synthetic training examples, and $\theta$ denotes the model parameters.

For this purpose, we use \textsc{Bart} \citep{lewis-etal-2020-bart} as the foundation for fine-tuning the corrector due to its proven effectiveness in conditional text generation tasks. 
\textsc{Bart} is a seq2seq auto-regressive transformer pre-trained on various denoising objectives, such as text infilling and token deletion, making it adept at recovering the original text from corrupted input. 
This pre-training aligns naturally with our summary correction task, where the model treats the incorrect summary 
as noisy input, focusing on resolving errors to recover factual consistency.

\section{Experiments}
\subsection{Dataset}
Our research utilizes \textsc{MentSum}, the first mental health summarization dataset, which contains selected user posts from Reddit along with their short user-written summaries (called TL;DR) in English. Each lengthy post articulates a user's mental health problem and quest for support from community and professional counselors, while the corresponding TL;DR serves to condense this narrative into a concise summary, facilitating quicker review and response by counselors. 
This dataset comprises over 24k post-TL;DR pairs, divided into 21,695 training, 1,209 validation, and 1,215 test instances. On average, each post contains 327.5 words or 16.9 sentences, while TL;DR consists of 43.5 words or 2.6 sentences. 
More details about the dataset can be found in \citet{sotudeh-etal-2022-mentsum}.

\subsection{Metrics}
To evaluate the linguistic quality of the generated summary, we use standard \textsc{Rouge} metrics: \textsc{Rouge-1}, \textsc{Rouge-2}, and \textsc{Rouge-L}. These metrics assess the overlap of unigrams, bigrams, and the longest common subsequence, respectively, between the generated summary and reference one. We report the F1 scores for these metrics to provide a comprehensive analysis.

For automatically assessing the factual consistency of the generated summary with the source document, we utilize a fine-tuned version of the FactCC model \citep{kryscinski-etal-2020-evaluating}. 
This model maps the consistency evaluation as a binary classification problem, and outputs a probability score ranging from 0 to 1, indicating the likelihood that the generated summary is factually consistent with the source content. 

\subsection{Implementation Details}
\paragraph{Guided Summarizer.} \label{details}
To construct knowledge-based guidance, we curate mental health terminologies from subsets released by Kaiser Permanente (KP) in 2011 and 2016\footnote{\url{https://www.johnsnowlabs.com/marketplace/cmt-mental-health-problem-list-subset/}}, focusing on the ``KP\_Patient\_Display\_Name” column. 
Our preprocessing involves (1) separating terms that are combined with commas to ensure each term is individually identifiable, (2) splitting terms that contain parentheses (e.g., ``A (B)”) into two separate entities to simplify and clarify the data, (3) removing duplicates to compile a list of unique terms, and (4) excluding terms longer than three words to improve regex matching efficiency. This process yields a refined list of 1,068 unique terminological terms.
Then, we extract these identified terms from each mental health post, separate them with a special [SEP] token, and use them as the first type of guidance.
Additionally, we explore an alternative approach by extracting sentences from each source post that contain any of the predefined terminology, using them as the second type of guidance.
Regular expressions are employed to ensure a precise match of the entire term, avoiding partial or irrelevant matches.

We adopt the \textsc{Bart}-large as the foundation for fine-tuning our guided summarization model\footnote{\url{https://github.com/neulab/guided\_summarization}}. 
Training parameters include a total of 10,000 updates, a maximum token of 1,024, and an update frequency of 4. We opt for the AdamW optimizer with a learning rate of 3e-5, $\beta$ parameters set to (0.9, 0.98), and a weight decay of 0.01. 
The objective function is cross-entropy Loss across all models. 
After training for five epochs, the model checkpoint achieving the highest \textsc{Rouge-L} score on the validation set is selected for inference. 
For decoding, we employ a beam size of 6, with minimum and maximum lengths set to 15 and 200, respectively, and a restriction on repeating trigrams. 
All our experiments are conducted on four NVIDIA Tesla V100 GPUs, with the training process requiring approximately four hours.

\begin{table*}[ht]
    \centering
    \small
    \begin{tabular}{llllll}
         \toprule
         \textbf{Model} & \textbf{Guidance Signal} & \textbf{\textsc{Rouge-1}} & \textbf{\textsc{Rouge-2}} & \textbf{\textsc{Rouge-L}} & \textbf{FactCC}  \\
         \midrule
         \textsc{CurrSum} & No signal & \textit{30.16} & \textit{8.82} & \textit{21.24} & -- \\
         \midrule
         \textsc{Bart} & \multirow{2}{*}{No signal} & 28.792 & 8.741 & 23.657 & 87.74 \\
         $\quad$ \textit{After Correction} && 28.754 & 8.722 & 23.625 & 88.40 (↑0.75\%) \\
         \midrule
         \textsc{GSum} & Highlighted & 30.031 & 8.917 & 24.698 & 87.65 \\
         $\quad$ \textit{After Correction} & sentences & 30.013 & 8.907 & 24.685 & 87.98 (↑0.38\%) \\
         \midrule
         \textbf{\textsc{GSum-term}} & Specialized & 30.429 & 9.441 & \textbf{25.335} & 89.05 \\
         $\quad$ \textit{After Correction} & terminologies & 30.426 & 9.425 & 25.326 & 89.22 (↑0.19\%)\\
         \midrule
         \textbf{\textsc{GSum-sent}} & Context-rich & \textbf{30.578} & \textbf{9.647} & 25.315 & 90.12 \\
         $\quad$ \textit{After Correction} & sentences & 30.561 & 9.638 & 25.309 & \textbf{90.62} (↑0.55\%) \\
         \bottomrule
    \end{tabular}
    \caption{\textsc{Rouge} scores and FactCC scores on \textsc{MentSum} test set.}
    \label{tab1}
\end{table*}

\paragraph{Error Corrector.}
We create synthetic incorrect summaries incorporating entity, number, date, and pronoun errors, resulting in 25,940 training and 1,416 validation examples.
Based on the \textsc{Bart}-large architecture implemented in fairseq\footnote{\url{https://github.com/pytorch/fairseq/blob/master/examples/bart}}, the neural corrector is fine-tuned with the parameter setting similar to the guided summarizer, except it is trained for 10 epochs to allow the model to adequately learn to identify and correct these subtle errors. 
During inference, the candidate summary generated from the previous guided summarizer is concatenated with its source post, and processed by the optimal checkpoint to produce the corrected summary for final evaluation.

\paragraph{FactCC Evaluator.}
We re-implement and fine-tune the FactCC model\footnote{\url{https://github.com/salesforce/factCC}}, tailoring it to better suit our domain-specific needs. 
The training data consist of both correct and incorrect examples: the former derives from clean reference summary (labeled as ``CORRECT”), while the latter uses the same synthetic data as the corrector (labeled as ``INCORRECT”), signifying inconsistent with the source post. 
Thus, we obtain 21,695 correct and 25,940 incorrect examples for training, with 1,209 correct and 1,416 incorrect examples for validation.
Based on the \textsc{Bert}-base model, we use the same hyper-parameters for training the original FactCC model over 10 epochs.
For inference, the corrected summary (defined as ``claim”) and its corresponding source post (defined as ``text”) are combined and fed into the optimally selected checkpoint (with the lowest Loss) to compute a probability score, quantitatively evaluating the alignment between claim and text. 

\subsection{Baselines}
\paragraph{\textsc{Bart}.}
It is a pre-trained SOTA model for summarization tasks, and demonstrated superior performance over various extractive and abstractive summarizers on \textsc{MentSum} dataset \citep{sotudeh-etal-2022-mentsum}. 
We re-employ \textsc{Bart} on this dataset as a baseline rather than simply copying the results because that study did not evaluate factual consistency, a key focus of our research for comparison. 
In this baseline experiment, training parameters match those of the guided summarizer, with the exception of setting the update frequency to 1.

\paragraph{\textsc{GSum}.}

We adopt \textsc{GSum} with highlighted sentences, the best-performing guidance signal, as our second baseline.
Highlighted sentences are identified as oracle sentences during training using a greedy search algorithm for maximum \textsc{Rouge} scores with reference summaries, but are extracted during inference by employing a pre-trained extractive summarizer, i.e., the best-performing BertExt checkpoint \citep{liu-lapata-2019-text}, due to unseen references summaries in the test set.
This baseline experiment proves more complex compared to our guided summarizer, as it requires an additional summarizer during inference, a limitation within the original framework.
In contrast, our guidance extraction (described in Section \ref{details}) relies on specialized terminologies, moving away from reliance on oracle sentences, and streamlining the process by unifying the training and inference phases. 


\begin{table*}[htbp]
    \centering
    \scalebox{0.8}{
    \begin{tabular}{m{21em}m{7em}m{13em}m{6em}}
         \Xhline{1.1pt}
         \textbf{User posts} & \textbf{Human-written} & \textbf{\textsc{GSum-sent} generated} & \textbf{Corrected} \\
         \hline
         ...  \textit{About 6- 8 months later ther \textbf{anxiety} was back} ... So back to the prozac I went. \textcolor{red!60!black}{I took it again for about 18 months} and then the cycle continued ... \textit{I finally went to see a psychiatrist and she suggested I go back on it again since my \textbf{anxiety} has gotten out of control again} ...  & going back on Prozac for the third round ... & I started taking prozac about 4 years ago, stopped taking it after 18 months and \textcolor{red!60!black}{then took it again for about 6 - 8 months}. Now I 'm back on it again and my \textbf{anxiety} has gotten out of control again ... & ... \textcolor{red!60!black}{then took it again for about 18 months} ... \\
         \hline
         \textit{I'm  19 year old male with \textbf{autism, depression and anxiety}} ... \textit{Since then my dad has sunk into a deep \textbf{depression}}... and he has threatened to kill himself ... \textcolor{red!60!black}{My mother who is also extremely sickly has dealt with him for years ... and is probably going to leave him.} On top of all this we cannot afford to continue to live in our house ... & mom and dad sick, dad threading suicide, mom leaving dad, we re losing our house ... & I m a 19 year old male with \textbf{autism, depression and anxiety} who ca nt afford to live in our house, my father is \textbf{depressed} and suicidal, \textcolor{red!60!black}{my mother is leaving me} ... & ... \textcolor{red!60!black}{my mother is leaving him} ...\\
         \hline
         ... \textcolor{red!60!black}{made an excellent suggestion w CBT, but if cost or availability is an issue, guidance is a great way to start} ... \textit{When I was first coming to grips with my \textbf{anxiety}, I journaled quite frequently. Not only was it a great \textbf{emotional} release to let my intrusive thoughts out} ... & I suggest looking into CBT or a guidance counselor. Journalling is good ... & \textcolor{red!60!black}{CBT is a great way to start}. Do n't be afraid to journal ... & \textcolor{red!60!black}{Guidance is a great way to start} ... \\
         \hline
         ... \textit{\textcolor{red!60!black}{I've been really \textbf{depressed} for years}} ... I'll spare you the details but this goes on for close to 10 years ... \textit{I've been failing everything I take due to a combination of \textbf{depression, anxiety} ...} & My life was messed up ... & \textcolor{red!60!black}{I 've been in and out of mental institutions for 10 years} and have been failing everything I take due to a combination of \textbf{depression, anxiety} ... & \textcolor{red!60!black}{I 've been in and out of mental institutions for years} ...\\
         \Xhline{1.1pt}
    \end{tabular}}
    \caption{Four samples of mental health posts along with their human-written TL;DRs, summaries generated by \textsc{GSum-sent}, and corrections applied by neural corrector.}
    \label{tab2}
\end{table*}

\section{Results}
\subsection{Automatic Evaluation}
We refer to the model that utilizes extracted mental health terminologies for guidance as \textsc{GSum-term}, and the model that employs context-rich sentences containing these domain-specific terms for guidance as \textsc{GSum-sent}. 
As shown in Table \ref{tab1}, experimental results on the \textsc{MentSum} dataset reveal that our proposed models surpass established benchmarks and baseline models in terms of both \textsc{Rouge} and FactCC scores. Below, we present more detailed insights derived from our experiments. 

\paragraph{Effectiveness of Guided Summarizer.}
Initial observations from two baseline experiments indicate that guided summarizer exhibits improved \textsc{Rouge} scores, particularly in the \textsc{Rouge-2} and \textsc{Rouge-L} metrics, compared to \textsc{CurrSum}, indicating a better capture of detailed information and narrative structure. 
However, the original \textsc{GSum} achieves a lower FactCC score compared to \textsc{Bart}, suggesting that while highlighted sentences can steer the model toward relevant information, they do not guarantee factual consistency. 

\paragraph{Improvement through Domain-specific Guidance.}
Our experiments with the proposed models yielded significant improvements on both \textsc{Rouge} and FactCC scores over the baseline models, indicating improvements in summary quality and factual consistency. 
Specifically, \textsc{GSum-term} is 1.5\% higher than \textsc{Bart} and 1.6\% higher than \textsc{GSum} on FactCC score, suggesting that the use of specialized terminologies as guidance signal, instead of highlighted sentences, is effective in enhancing the summary's alignment with the source content while maintaining or even improving its overall quality. 

The subsequent experiment with the \textsc{GSum-sent} model employs context-rich sentences embedded with domain-specific terms as the guidance signal, leading to notable advancement across the board.
Specifically, the model not only records superior \textsc{Rouge} scores but also achieves a 2.7\% higher FactCC score compared to \textsc{Bart} and 2.8\% improvement over \textsc{GSum}. 
This finding, resonating with the insight from the original \textsc{GSum} study, highlights the superiority of contextually rich, sentence-based guidance over simpler keyword-based one.
Overall, this integration of domain-specific guidance underscores the importance of leveraging specialized information from the source post, and is pivotal for the generated summary to improve its alignment with the source content in the mental health context. 

\paragraph{Benefit of Corrector.}
The correction model demonstrates its capability to refine the consistency of summary and faithfully represent the source details across both our proposed models and baseline models.
After correction, the FactCC score sees an uplift ranging from 0.19\% to 0.75\% across all models evaluated. 
It’s worth noticing that correction generally results in a slight decline in \textsc{Rouge} scores, a phenomenon observed in multiple studies \citep{kryscinski-etal-2020-evaluating, maynez-etal-2020-faithfulness}, and may be attributed to the nuanced balance between enhancing 
factual consistency and maintaining linguistic quality in the summary.

\subsection{Case Study and Analysis}
Acknowledging the limitations of automatic evaluation in the summarization system, we also manually assess the quality of our work by comparing candidate summaries generated by \textsc{GSum-sent} and corrected ones against human-written TL;DRs, as shown in Table \ref{tab2}.
To protect user privacy, the source posts are selectively displayed. The specialized mental health terminologies are highlighted in \textbf{bold}, and sentences containing these terms are in \textit{italic} to show their influence on the summary generation process. Additionally, corrections and related text segments are marked in \textcolor{red!60!black}{red} to provide clear insight into the improvements in detail consistency.

\paragraph{Heightened Domain Specificity.} Summaries generated by \textsc{GSum-sent} often capture more specialized mental health terms.
Conversely, TL;DRs are written in a colloquial and condensed manner, which might omit essential terminological details. 
Taking the first sample as an example, the human-written summary merely mentions going back on Prozac for the third time, while the \textsc{GSum-sent}-generated one specifies details on the duration of treatment and the underlying issue of anxiety.
Similarly, in the fourth sample, the human-written summary describes the situation as ``messed up”, a vague term compared to the explicit mentions of ``depression” and ``anxiety” by \textsc{GSum-sent}.
They all indicate the model's potential to provide more transparent communication of mental health issues, which is helpful when asking for support from professional counselors. 

\paragraph{Improved Faithfulness.} 
Both the guided summarizer and corrector play crucial roles in improving faithfulness according to reported FactCC scores, with the corrector further enhancing detail consistency with respect to the source post.
It addresses date inaccuracy in the first and fourth samples, corrects pronoun usage in the second, and resolves entity error in the third. These errors originate from incorrect references to similar items within the original posts, exemplified by the misrepresentation of ``6-8 months” in the first sample. 

Despite the precision in correction, there is a shortcoming: the corrector's modifications are very subtle, attributed to its training on a dataset limited to four types of minor errors. 
This restraint in correction is evident in our examination of summaries generated by \textsc{GSum-sent} model, where only 10.3\% undergo revisions by corrector. Moreover, these adjustments are minimal, with 92.8\% of the corrected summaries incorporating three or fewer new tokens, despite the summary averaging 53.27 tokens in length. 
This indicates that the current correction model may not fully capture complex inaccuracies beyond its training scope, highlighting the need for a more diverse training dataset to enhance its ability to improve detail consistency across a wider range of summaries.

\section{Conclusion}
Focusing on the mental health domain, our research addresses the challenges of generating domain-relevant and faithful summaries through the development of a guided summarizer followed by a neural corrector. 
By incorporating novel domain-specific knowledge-based guidance, especially context-rich sentences, our adapted summarizer closely aligns with the specialized source content and effectively enhances the domain relevance of the generated summary. 
The post-editing corrector further ensures the elimination of inconsistency or intrinsic hallucination, making the summary more faithful to the source document. 

Comprehensive evaluation with the \textsc{MentSum} dataset demonstrates the superior performance of our proposed model over existing baselines, as evidenced by improvements in both \textsc{Rouge} and FactCC scores.
Although our experiments are specifically tailored to the mental health domain, 
the methodologies we've developed are adaptable across various fields where the precision of domain-specific knowledge and detail consistency are both essential, such as in legal, financial, or technical contexts. 
The effectiveness and adaptability of our approach underscore its potential to advance domain-specific abstractive summarization, offering a versatile framework for future exploration.

\section*{Acknowledgments}
We thank to anonymous reviewers for their valuable comments, and Georgetown University for providing the \textsc{MentSum} dataset.
This work is partially supported by the 2022 Jiangsu Science and Technology Program (General Program), contract number BK20221260.



\bibliography{anthology, custom}

\appendix



\end{document}